\documentclass{article}

\usepackage[nonatbib]{nips_2017}

\usepackage{nips_2017}

\usepackage[utf8]{inputenc} 
\usepackage[T1]{fontenc}    
\usepackage{hyperref}       
\usepackage{url}            
\usepackage{booktabs}       
\usepackage{amsfonts}       
\usepackage{nicefrac}       
\usepackage{microtype}      

\usepackage{graphicx} 
\usepackage{subfigure} 
\usepackage{amsmath}
\DeclareMathOperator{\Tr}{Tr}
\DeclareMathOperator*{\argmin}{argmin} 
\usepackage{tabularx}
\usepackage{multirow}
\usepackage[ruled,vlined]{algorithm2e}

\usepackage{booktabs}
\newcommand{\ra}[1]{\renewcommand{\arraystretch}{#1}}

\usepackage{lipsum}

\usepackage{wrapfig}

\title{GAR: An efficient and scalable Graph-based Activity Regularization for semi-supervised learning}

\author{
  Ozsel~Kilinc \\
  Electrical Engineering Department\\
  University of South Florida\\
  Tampa, FL 33620 \\
  \texttt{ozsel@mail.usf.edu} \\
  \And
  Ismail~Uysal  \\
  Electrical Engineering Department \\
  University of South Florida \\
  Tampa, FL 33620  \\
  \texttt{iuysal@usf.edu} \\
}

\begin{document}

\maketitle

\begin{abstract}
In this paper, we propose a novel graph-based approach for semi-supervised learning problems, which considers an adaptive adjacency of the examples throughout the unsupervised portion of the training. Adjacency of the examples is inferred using the predictions of a neural network model which is first initialized by a supervised pretraining. These predictions are then updated according to a novel unsupervised objective which regularizes another adjacency, now linking the output nodes. Regularizing the adjacency of the output nodes, inferred from the predictions of the network, creates an easier optimization problem and ultimately provides that the predictions of the network turn into the optimal embedding. Ultimately, the proposed framework provides an effective and scalable graph-based solution which is natural to the operational mechanism of deep neural networks.  Our results show comparable performance with state-of-the-art generative approaches for semi-supervised learning on an easier-to-train, low-cost framework.

\end{abstract}

\section{Introduction}

The idea of utilizing an auxiliary unsupervised task to help supervised learning dates back to 90s \cite{suddarth90}. As an example to one of its numerous achievements, unsupervised pretraining followed by supervised fine-tuning was the first method to succeed in the training of fully connected architectures \cite{hinton2006fast}. Although today it is known that unsupervised pretraining is not a must for successful training, this particular accomplishment played an important role enabling the current deep learning renaissance and has become a canonical example of how a learning representation for one task can be useful for another one \cite{Goodfellow-et-al-2016}. There exist a variety of approaches to combine supervised and unsupervised learning in the literature. More specifically, the term \textit{semi-supervised} is commonly used to describe a particular type of learning for applications in which there exists a large number of observations, where a small subset of them has ground-truth labels. Proposed approaches aim to leverage the unlabeled data to improve the generalization of the learned model and ultimately obtain a better classification performance.

One approach is to introduce additional penalization to the training based on the reconstruction of the input through autoencoders or their variants \cite{RanzatoS08}. Recently, there have been significant improvements in this field following the introduction of a different generative modeling technique which uses the variational equivalent of deep autoencoders integrating stochastic latent variables into the conventional architecture \cite{KingmaW13} \cite{RezendeMW14}. First, \cite{KingmaMRW14} have shown that such modifications make generative approaches highly competitive for semi-supervised learning. Later, \cite{MaaloeSSW16} further improved the results obtained using variational autoencoders by introducing auxiliary variables increasing the flexibility of the model. Furthermore, \cite{RasmusBHVR15} have applied Ladder networks \cite{Valpola14}, a layer-wise denoising autoencoder with skip connections from the encoder to the decoder, for semi-supervised classification tasks. These recent improvements have also motivated researchers to offer radically novel solutions such as virtual adversarial training \cite{MiyatoMKNI15} motivated by Generative Adversarial Nets \cite{GoodfellowPMXWOCB14} proposing a new framework corresponding to a minimax two-player game. Alternatively, \cite{YangCS16} revisited graph-based methods with new perspectives such as invoking the embeddings to predict the context in the graph. 

Conventional graph-based methods aim to construct a graph propagating the label information from labeled to unlabeled observations and connecting similar ones using a graph Laplacian regularization in which the key assumption is that nearby nodes are likely to have the same labels. \cite{ZhuGL03} and \cite{BelkinN03} are examples of transductive graph-based methods suffering from the out-of-sample problem, i.e. they can only predict examples that are already observed in the training set. \cite{HeYHNZ05}, \cite{BelkinNS06}, \cite{NieXTZ10} and \cite{WangNHCYY17} addressed this problem and proposed inductive frameworks where predictions can be generalized to never-seen-before examples. In a different work, \cite{WestonRMC12} have shown that the idea of combining an embedding-based regularizer with a supervised learner to perform semi-supervised learning can be generalized to deep neural networks and the resulting inductive models can be trained by stochastic gradient descent. A possible bottleneck with this approach is that the optimization of the unsupervised part of the loss function requires precomputation of the weight matrix specifying the similarity or dissimilarity between the examples whose size grows quadratically with the number of examples. For example, a common approach to computing the similarity matrix is to use the $k$-nearest neighbor algorithm which must be approximated using sampling techniques as it becomes computationally very expensive for a large number of observations. Both the graph-construction (due to the adopted auxiliary algorithm such as nearest neighbor) and the regularization (due to the requirement for eigenanalysis of the graph Laplacian) processes severely limit the scalability of conventional graph-based methods. \cite{LiuHC10} and \cite{WangFHTW16} addressed the scalability issue of graph-based methods using a small number of anchor points adequately covering the data distribution. Hierarchical Anchor Graph Regularization (HAGR) \cite{WangFHLW17} further improved the scalability of anchor-graph-based approaches by exploring multiple-layer anchors with a pyramid-style structure.

In this paper, we propose a novel framework for semi-supervised learning which can be considered a variant of graph-based approach. This framework can be described as follows.

\begin{itemize}
	\item Instead of a graph between examples, we consider a bipartite graph between examples and output classes. To define this graph, we use the predictions of a neural network model initialized by a supervised pretraining which uses a small subset of samples with known labels.
	\item This bipartite graph also infers two disjoint graphs: One between the examples and another between the output nodes. We introduce two regularization terms for the graph between the output nodes and during the unsupervised portion of training, the predictions of the network are updated only based on these two regularizers. 
	\item These terms implicitly provide that the bipartite graph between the examples and the output classes becomes a biregular graph and the inferred graph between the examples becomes a disconnected graph of regular subgraphs. Ultimately, because of this observation, the predictions of the network yield the embeddings that we try to find.
\end{itemize}

The proposed framework is naturally inductive and more importantly, it is scalable and it can be applied to datasets regardless of the sample size or the dimensionality of the feature set. Furthermore, the entire framework operationally implements the same feedforward and backpropagation mechanisms of the state of the art deep neural networks as the proposed regularization terms are added to the loss function as naturally as adding standard $L1$, $L2$ regularizations \cite{ng04l1l2} and similarly optimized using stochastic gradient descent \cite{bottou10sgd}. 

\section{Related work}

Consider a semi-supervised learning problem where out of $m$ observations, corresponding ground-truth labels are only known for a subset of $m_L$ examples and the labels of complimentary subset of $m_U$ examples are unknown where typically $m_L\ll m_U$. Let $\boldsymbol{x}_{1:m}$ and $y_{1:m}$ denote the input feature vectors and the output predictions respectively and $t_{1:m_L}$ denote the available output labels. The main objective is to train a classifier $f: \boldsymbol{x} \rightarrow y$ using all $m$ observations that is more accurate than another classifier trained using only the labeled examples. Graph-based semi-supervised methods consider a connected graph $\mathcal{G}=(\mathcal{V},\mathcal{E})$ of which vertices $\mathcal{V}$ correspond to all $m$ examples and edges $\mathcal{E}$ are specified by an $m \times m$ adjacency matrix $\boldsymbol{A}$ whose entries indicate the similarity between the vertices. There have been many different approaches about the estimation of the adjacency matrix $\boldsymbol{A}$. \cite{ZhuGL03} derive $\boldsymbol{A}$ according to simple Euclidean distances between the samples while \cite{WestonRMC12} precompute $\boldsymbol{A}$ using $k$-nearest neighbor algorithm. They also suggest that, in case of a sequential data, one can presume consecutive instances are also neighbors on the graph. \cite{YangCS16}, on the other hand, consider the specific case where $\boldsymbol{A}$ is explicitly given and represents additional information. The most important common factor in all these graph-based methods is the fact that $\boldsymbol{A}$ is a fixed matrix throughout the training procedure with the key assumption that nearby samples on $\mathcal{G}$, which is defined by $\boldsymbol{A}$, are likely to have the same labels. Hence, the generic form of the loss function for these approaches can be written as:
\begin{equation}
\sum\limits_{i =1}^{m_L}{\mathcal{L}\big(f(\boldsymbol{x}_i), t_i\big)} + 
\lambda\sum\limits_{i,j=1}^{m}{\mathcal{U}\big(f(\boldsymbol{x}_i), f(\boldsymbol{x}_j), A_{ij}\big)}
\end{equation}
where $\mathcal{L}(.)$ is the supervised loss function such as log loss, hinge loss or squared loss, $\mathcal{U}(.)$ is the unsupervised regularization (in which a multi-dimensional embedding $g(\boldsymbol{x}_i) = \boldsymbol{z}_i$ can also be replaced with one-dimensional $f(\boldsymbol{x}_i)$) and $\lambda$ is a weighting coefficient between supervised and unsupervised metrics of the training $\mathcal{L}(.)$ and $\mathcal{U}(.)$. One of the commonly employed embedding algorithms in semi-supervised learning is Laplacian Eigenmaps \cite{BelkinN03} which describes the distance between the samples in terms of the Laplacian $\boldsymbol{L} = \boldsymbol{D} - \boldsymbol{A}$, where $\boldsymbol{D}$ is the diagonal degree matrix such that $D_{ii}=\sum_{j}A_{ij}$. Then, unsupervised regularization becomes:
\begin{equation}
\label{eigenmaps_loss}
\sum\limits_{i,j=1}^{m}{\mathcal{U}\big(g(\boldsymbol{x}_i), g(\boldsymbol{x}_j), A_{ij}\big)} = \sum\limits_{i,j=1}^{m}{A_{ij}||g(\boldsymbol{x}_i) - g(\boldsymbol{x}_j)||^2} = \Tr(\boldsymbol{Z}^T\boldsymbol{LZ})
\end{equation}
subject to the balancing constraint $\boldsymbol{Z}^T\boldsymbol{DZ}=\boldsymbol{I}$, where $\boldsymbol{Z} = [\boldsymbol{z}_1,...,\boldsymbol{z}_m]^T$. \cite{ZhuGL03} use this regularization together with a nearest neighbor classifier while \cite{BelkinNS06} integrate hinge loss to train an SVM. Both methods impose regularization on labels $f(\boldsymbol{x}_i)$. 
On the other hand, \cite{WestonRMC12} employ a margin-based regularization by \cite{HadsellCL06} such that 
\begin{equation}
\mathcal{U}\big(f(\boldsymbol{x}_i), f(\boldsymbol{x}_j), A_{ij}\big) =
\begin{cases}
||f(\boldsymbol{x}_i) - f(\boldsymbol{x}_j)||^2 & \text{if $A_{ij} = 1$} \\ 
max(0, \gamma-||f(\boldsymbol{x}_i) - f(\boldsymbol{x}_j)||^2) & \text{if $A_{ij} = 0$}
\end{cases}
\end{equation}
to eliminate the balancing constraints and enable optimization using gradient descent. They also propose to learn multi-dimensional embeddings on neural networks such that $g(\boldsymbol{x}_i)=f^l(\boldsymbol{x}_i) = \boldsymbol{y}^l_i$, where $\boldsymbol{y}^l_i$ is the output of the $l$\textsuperscript{th} hidden layer corresponding to the $i$\textsuperscript{th} sample.

\section{Proposed framework}

\subsection{Bipartite graph approach}

Instead of estimating the adjacency matrix $\boldsymbol{A}$ using an auxiliary algorithm such as nearest neighbor or auxiliary external knowledge, we propose to use the actual predictions of a neural network model initialized by a supervised pretraining using the subset of $m_L$ labeled examples. 

Suppose that after pretraining, predictions of the network, $\boldsymbol{B}$, for all $m$ examples are obtained as an $m \times n$ matrix, where $n$ is the number of output classes and $B_{ij}$ is the probability of the $i$\textsuperscript{th} example belonging to $j$\textsuperscript{th} class. We observe that these predictions, indeed, define a bipartite graph $\mathcal{G}^*=(\mathcal{V}^*,\mathcal{E}^*)$ whose vertices $\mathcal{V}^*$ are $m$ examples together with $n$ output nodes.  However, $\mathcal{V}^*$  can be divided into two disjoint sets $\mathcal{M}$ and $\mathcal{N}$, such that $\mathcal{G}^*=(\mathcal{M},\mathcal{N},\mathcal{E}^*)$, where $\mathcal{M}$ is the set of examples, $\mathcal{N}$ is the set of output nodes and an edge $\mathsf{e} \in \mathcal{E}^*$ connects an example $\mathsf{m} \in \mathcal{M}$ with an output node $\mathsf{n} \in \mathcal{N}$. As there is no lateral connection between $m$ examples and between $n$ output nodes, ($m + n) \times (m + n$) adjacency matrix $\boldsymbol{A}^*$ of graph $\mathcal{G}^*$ has the following form
\begin{equation}
\boldsymbol{A}^*=
\begin{pmatrix}
\boldsymbol{0}_{m \times m} & \boldsymbol{B}_{m \times n}  \\
\boldsymbol{B}^T_{n \times m} & \boldsymbol{0}_{n \times n}
\end{pmatrix}
\end{equation}
where $\boldsymbol{B}$ corresponds to $m \times n$ biadjacency matrix of graph $\mathcal{G}^*$ which is by itself unique and sufficient to describe the entire $\mathcal{E}^*$. 

In graph $\mathcal{G}^*$, the examples are connected with each other by even-length walks through the output nodes whereas the same is true for the output nodes through the samples. 
In this case, the square of the adjacency matrix $\boldsymbol{A}^*$ represents the collection of two-walks (walks with two edges) between the vertices. It also implements two disjoint graphs $\mathcal{G}_\mathcal{M}=(\mathcal{M}, \mathcal{E}_\mathcal{M})$ and $\mathcal{G}_\mathcal{N}=(\mathcal{N}, \mathcal{E}_\mathcal{N})$ such that
\begin{equation}
{\boldsymbol{A}^*}^2=
\begin{pmatrix}
\boldsymbol{B}\boldsymbol{B}^T_{m \times m} & \boldsymbol{0}_{m \times n}  \\
\boldsymbol{0}_{n \times m} & \boldsymbol{B}^T\boldsymbol{B}_{n \times n} 
\end{pmatrix} =
\begin{pmatrix}
\boldsymbol{M} & \boldsymbol{0}  \\
\boldsymbol{0} & \boldsymbol{N}
\end{pmatrix}
\end{equation}
where $\boldsymbol{M}=\boldsymbol{BB}^T$ and $\boldsymbol{N}=\boldsymbol{B}^T\boldsymbol{B}$ are the adjacency matrices specifying edges $\mathcal{E}_\mathcal{M}$ and $\mathcal{E}_\mathcal{N}$, respectively. Unlike a simple graph $\mathcal{G}$ considered by conventional graph-based methods, $\mathcal{G}_\mathcal{M}$ also involves self-loops. However, they have no effect on the graph Laplacian and thus on the embeddings. Hence, one can estimate the adjacency of examples using the predictions of the pretrained neural network, i.e. $\boldsymbol{M}=\boldsymbol{B}\boldsymbol{B}^T$, and then find the embeddings by applying a standard unsupervised objective such as Laplacian Eigenmap minimizing $\Tr(\boldsymbol{Z}^T\boldsymbol{L}_\mathcal{M}\boldsymbol{Z})$ as defined in (\ref{eigenmaps_loss}), where $\boldsymbol{L}_\mathcal{M} = \boldsymbol{D}_\mathcal{M} - \boldsymbol{M}$. It is important to note that conventional graph-based algorithms assume a fixed adjacency matrix during loss minimization whereas in the proposed framework, we consider an adaptive adjacency which is updated throughout the unsupervised training process as described in the following section.

\subsection{Adaptive adjacency}
\label{sec:adaptive_adjacency}

As derived adjacency $\boldsymbol{M}$ depends only on $\boldsymbol{B}$, during the unsupervised training, 
$\boldsymbol{B}$ needs to be well-constrained to preserve the learned latent embeddings. Otherwise, the idea of updating the adjacency throughout an unsupervised task might be catastrophic and results in offsetting the effects of the supervised pretraining. 

There are two constraints derived from the natural expectation on the specific form of the $\boldsymbol{B}$ matrix for a classification problem: i) first, a sample is to be assigned to one class with the probability of 1, while remaining $n-1$ classes have 0 association probability, and ii) second, for balanced classification tasks, each class is expected to involve approximately the same number of the examples. Let us first consider the perfect case, where $\boldsymbol{B}$ assigns $\nicefrac{m}{n}$ examples to each one of the $n$ classes with the probability of 1. $\boldsymbol{B}$ in this particular form implies that graph $\mathcal{G}^*$ becomes $(1,\nicefrac{m}{n})$-biregular since 
\begin{equation}
\deg(\mathsf{m}_i)=1, \forall \mathsf{m}_i \in \mathcal{M} 
\textnormal{\space\space\space  and \space\space\space } 
\deg(\mathsf{n}_i)=\nicefrac{m}{n}, \forall \mathsf{n}_i \in \mathcal{N}
\end{equation}

Subsequently graph $\mathcal{G}_\mathcal{N}$ turns into a disconnected graph including only self-loops and its adjacency matrix $\boldsymbol{N}$ becomes a scaled identity matrix indicating mutually exclusive and uniform distribution of the examples across the output nodes. Similarly, $\mathcal{G}_\mathcal{M}$ also becomes a disconnected graph including $n$ disjoint subgraphs. Each one of these subgraphs is $\nicefrac{m}{n}$-regular, where each vertex also has an additional self-loop. Hence, the degree matrix $\boldsymbol{D}_\mathcal{M}$ becomes a scaled identity matrix and $\boldsymbol{M}$ yields that $\boldsymbol{B}$ represents the optimal embedding. As they all depend only on $\boldsymbol{B}$, the relationships between $\mathcal{G}^*$, $\mathcal{G}_\mathcal{M}$ and $\mathcal{G}_\mathcal{N}$ are biconditional, which can be written as follows:
\begin{equation}
\label{relation}
\mathcal{G}^* \textnormal{ is } (1,\nicefrac{m}{n})\textnormal{-biregular} \Leftrightarrow
\mathcal{G}_\mathcal{M}:\argmin_{\substack{
											\boldsymbol{Z} \\
											\boldsymbol{Z}^T\boldsymbol{D}_\mathcal{M}\boldsymbol{Z}=\boldsymbol{I}}}{\Tr(\boldsymbol{Z}^T\boldsymbol{L}_\mathcal{M}\boldsymbol{Z}})=\boldsymbol{B} \Leftrightarrow
\mathcal{G}_\mathcal{N}:\boldsymbol{N} = \nicefrac{m}{n}\boldsymbol{I}
\end{equation}

Depending on this relation, we propose to apply regularization during the unsupervised task in order to ensure that $\boldsymbol{N}$ becomes the identity matrix. Regularizing $\boldsymbol{N}$ enables us to devise a framework that is natural to the operation of neural networks and compatible with stochastic gradient descent. The first one of the proposed two regularization terms constrains $\boldsymbol{N}$ to be a diagonal matrix, whereas the second one forces it into becoming a scaled identity matrix by equalizing the value of the diagonal entries. The second term ultimately corresponds to constraining $\boldsymbol{B}$ to obtain a uniform distribution of samples across the output classes. Obviously, this condition is not valid for every dataset, but a balancing constraint is required to prevent collapsing onto a subspace of dimension less than $n$ and it is analogous to the constraint of $\boldsymbol{Z}^T\boldsymbol{DZ}=\boldsymbol{I}$ in \cite{BelkinN03}. As the relation in (\ref{relation}) implies, when $\boldsymbol{N}$ becomes the identity matrix, $\boldsymbol{B}$ automatically represents the optimal embedding found using the inferred adjacency $\boldsymbol{M}$. In other words, no additional step is necessary to find the embeddings of $\mathcal{G}_\mathcal{M}$, as $\boldsymbol{B}$ already yields them.

\subsection{Activity regularization}
   
Consider a neural network with $L-1$ hidden layers where $l$ denotes the individual index for each hidden layer such that $l \in \{1,...,L\}$. Let $\boldsymbol{Y} ^{(l)}$ denote the output of the nodes at layer $l$. $\boldsymbol{Y} ^{(0)}=\boldsymbol{X}$ is the input and $f(\boldsymbol{X})=f^{(L)}(\boldsymbol{X})=\boldsymbol{Y}^{(L)}=\boldsymbol{Y}$ is the output of the entire network. $\boldsymbol{W} ^{(l)}$ and $\textbf{b}^{(l)}$ are the weights and biases of layer $l$, respectively. Then, the feedforward operation of the neural networks can be written as 
\begin{equation}
\boldsymbol{Y}^{(l)} = f^{(l)}\big(\boldsymbol{X}\big) = h^{(l)}\big(\boldsymbol{Y}^{(l-1)}\boldsymbol{W}^{(l)} + \boldsymbol{b}^{(l)}\big)
\end{equation}
where $h^{(l)}$(.) is the activation function applied at layer $l$.

In the proposed framework, instead of using the output probabilities of the softmax nodes, we use the activations at their inputs to calculate the regularization. The intuition here is that regularizing linear activations rather than nonlinear probabilities defines an easier optimization task. Since the multiplication of two negative activations yields a positive (false) adjacency in $\boldsymbol{M}$, we rectify the activations first. Then, $\boldsymbol{B}$ becomes
\begin{equation}
\boldsymbol{B} = g\big(\boldsymbol{X}\big) = \max{\bigg(\boldsymbol{0}, \big(\boldsymbol{Y}^{(L-1)}\boldsymbol{W}^{(L)} + \boldsymbol{b}^{(L)}\big)\bigg)}
\end{equation}

Recall that $\boldsymbol{N}$ is a $n \times n$ symmetric matrix such that $\boldsymbol{N}:=\boldsymbol{B}^T\boldsymbol{B}$ 
and let $\boldsymbol{v}$ be a $1 \times n$ vector representing the diagonal entries of $\boldsymbol{N}$ such that $\boldsymbol{v}:=[N_{11} \dots N_{nn}]$. Then, let us define $\boldsymbol{V}$ as a $n \times n$ symmetric matrix such that $\boldsymbol{V}:=\boldsymbol{v}^T\boldsymbol{v}$.
Then, the two proposed regularization terms can be written as

\begin{minipage}{0.5\linewidth}
\begin{equation}
\label{affinity}
\textit{Affinity} = \alpha\big(\boldsymbol{B}\big) :=\frac{\sum\limits_{i \ne j}^n{N_{ij}}}{(n-1)\sum\limits_{i = j}^n{N_{ij}}}
\end{equation}
\end{minipage}%
\begin{minipage}{0.5\linewidth}
\begin{equation}
\label{balance}
\textit{Balance} = \beta\big(\boldsymbol{B}\big) := \frac{\sum\limits_{i \ne j}^n{V_{ij}}}{(n-1)\sum\limits_{i = j}^n{V_{ij}}}
\end{equation}
\end{minipage}

While \textit{affinity} penalizes the non-zero off-diagonal entries of $\boldsymbol{N}$, \textit{balance} attempts to equalize diagonal entries. One might suggest minimizing the off-diagonal entries of $\boldsymbol{N}$ directly without normalizing, however, normalization is required to bring both regularizers within the same range for optimization and ultimately to ease hyperparameter adjustment. Unlike regularizing $\boldsymbol{N}$ to simply become a diagonal matrix, equalizing the diagonal entries is not an objective that we can reach directly by minimizing some entries of $\boldsymbol{N}$. Hence, we propose to use (\ref{balance}) that takes values between 0 and 1 where 1 represents the case where all diagonal entries of $\boldsymbol{N}$ are equal. Respectively, we propose to minimize the normalized term (\ref{affinity}) instead of the direct summation of the off-diagonal entries. However, during the optimization, denominators of these terms increase with the activations which may significantly diminish the effects of both regularizers. To prevent this phenomenon, we apply $L^2$ norm to penalize the overall activity increase. Recall that Frobenius norm for $\boldsymbol{B}$, $||\boldsymbol{B}||_F$ is analogous to the $L^2$ norm of a vector. Hence, the proposed overall unsupervised regularization loss ultimately becomes
\begin{equation}
\label{unsupervised_objective}
\mathcal{U}\big(g\big(\boldsymbol{X}\big)\big) = \mathcal{U}\big(\boldsymbol{B}\big)= c_{\alpha}\alpha\big(\boldsymbol{B}\big) + c_{\beta}\big(1-\beta\big(\boldsymbol{B}\big)\big) + c_F||\boldsymbol{B}||^2_F
\end{equation}

\subsection{Training}

Training of the proposed framework consists of two sequential steps: Supervised pretraining and subsequent unsupervised regularization. 
We adopt stochastic gradient descent in the mini-batch mode \cite{bottou10sgd} for optimization of both steps. Indeed, mini-batch mode is required for the unsupervised task since the proposed regularizers implicitly depend on the comparison of the examples with each other. Algorithm 1 below describes the entire training procedure. Pretraining is a typical supervised training task in which $m_L$ examples $\boldsymbol{X}_L=[\boldsymbol{x}_1 \dots \boldsymbol{x}_{m_L}]^T$ are introduced to the network with the corresponding ground-truth labels $\boldsymbol{t}_L=[{t}_1 \dots {t}_{m_L}]^T$ and the network parameters are updated to minimize the log loss $\mathcal{L}(.)$. After the pretraining is completed, this supervised objective is never revisited and the labels $\boldsymbol{t}_L$ are never reintroduced to the network in any part of the unsupervised task. Hence, the remaining unsupervised objective is driven only by the proposed regularization loss $\mathcal{U}(.)$ defined in (\ref{unsupervised_objective}). The examples $\boldsymbol{X}_L$ used in pretraining stage can also be batched together with the upcoming unlabeled examples $\boldsymbol{X}_U=[\boldsymbol{x}_{m_L+1} \dots \boldsymbol{x}_{m}]^T$ in order to ensure a more stable regularization. As the network is already introduced to them, $b_L$ examples randomly chosen from $\boldsymbol{X}_L$ can be used as guidance samples for the remaining $b_U$ examples of $\boldsymbol{X}_U$ in that batch. Such blended batches help the unsupervised task especially when the examples in the dataset have higher variance.   

\begin{algorithm}[t]
	\begin{small}
	\DontPrintSemicolon
	\SetKwFunction{proc}{proc}
	\SetKwProg{supervised}{Supervised pretraining:}{}{}
	\SetKwProg{unsupervised}{Unsupervised training:}{}{}
	\SetKwInOut{Input}{Input}
	
	\supervised{}{
		\Input{$\boldsymbol{X}_L=[\boldsymbol{x}_1 \dots \boldsymbol{x}_{m_L}]^T$,
			$\boldsymbol{t}_L=[{t}_1 \dots {t}_{m_L}]^T$, batch size $b$}
		\Repeat{\textnormal{stopping criteria is met}}
		{
			$ \big\{ (\boldsymbol{\acute{X}}_1, \boldsymbol{\acute{t}}_1),..., (\boldsymbol{\acute{X}}_{\nicefrac{m_L}{b}}, \boldsymbol{\acute{t}}_{\nicefrac{m_L}{b}})\big\} \longleftarrow (\boldsymbol{X}_L,\boldsymbol{t}_L)$ 
			\tcp*{Shuffle and create batch pairs}
			\For{$i \gets 1$ \textbf{to} $\nicefrac{m_L}{b}$}{
				Take $i$\textsuperscript{th} pair $(\boldsymbol{\acute{X}}_i, \boldsymbol{\acute{t}}_i)$ \\
				Take a gradient step for $\mathcal{L}\big(f\big(\boldsymbol{\acute{X}}_i\big),\boldsymbol{\acute{t}}_i\big)$ 
			}
		}
		\Return $model$ \\
	}
	
	\unsupervised{}{
		\SetKwFunction{random}{random}
		\Input{$model$, $\boldsymbol{X}_L=[\boldsymbol{x}_1 \dots \boldsymbol{x}_{m_L}]^T$, $\boldsymbol{X}_U=[\boldsymbol{x}_{m_L+1} \dots \boldsymbol{x}_{m}]^T$, $b_L$, $b_U$}
		\Repeat{\textnormal{stopping criteria is met}}
		{
			$ \big\{ \boldsymbol{\acute{X}}_1,..., \boldsymbol{\acute{X}}_{\nicefrac{m_U}{b_U}}, \big\} \longleftarrow \boldsymbol{X}_U$ \tcp*{Shuffle and create input batches} 
			\For{$i \gets 1$ \textbf{to} $\nicefrac{m_U}{b_U}$}{
				Take $i$\textsuperscript{th} input batch $\boldsymbol{\acute{X}}_i$ \\
				$\boldsymbol{\ddot{X}} \longleftarrow \random(\boldsymbol{X}_L, b_L)$ \tcp*{Randomly sample $b_L$ examples from $\boldsymbol{X}_L$}
				Take a gradient step for $\mathcal{U}\big(g\big(\begin{bmatrix}
				\boldsymbol{\acute{X}}_i^T &
				\boldsymbol{\ddot{X}}^T
				\end{bmatrix}^T\big)\big)$ 
			}
		}
	}
	\caption{Model training}
	
\end{small}
\end{algorithm}

\section{Experimental results}

The models have been implemented in Python using Keras \cite{chollet2015keras} and Theano \cite{Theano}. Open source code is available at \hyperref{http://github.com/ozcell/LALNets}{}{}{http://github.com/ozcell/LALNets} that can be used to reproduce the experimental results obtained on the three image datasets, MNIST \cite{lecun1998mnist}, SVHN \cite{svhn} and NORB \cite{norb} most commonly used by previous researchers publishing in the field of semi-supervised learning at NIPS and other similar venues. 

All experiments have been performed on convolutional neural network (CNN) models. A 6-layer CNN was used for MNIST, whereas for SVHN and NORB, a 9-layer CNN has been used. Coefficients of the proposed regularization term have been chosen as $c_\alpha=3, c_\beta=1$ and $c_F=0.000001$ in all of the experiments. We used a batch size of 128 for both supervised pretraining and unsupervised regularization steps. For each dataset, the same strategy is applied to decide on the ratio of labeled and unlabeled data in the unsupervised regularization batches, i.e. $\nicefrac{b_L}{b_U}$: $b_L$ is approximately assigned as one tenth of the number of all labeled examples $m_L$, i.e. $b_L \approx \nicefrac{m_L}{10}$, and then $b_U$ is chosen to complement the batch size up to 128, i.e. $b_U=128-b_L$. Each experiment has been repeated for 10 times. For each repetition, to assign $\boldsymbol{X}_L$, $m_L$ examples have been chosen randomly but distributed evenly across classes. Also, a validation set of 1000 examples has been chosen randomly among the training set examples to determine the epoch to report the test accuracy as is standard. 

\subsection{MNIST}

In MNIST, experiments have been performed using 4 different settings for the number of labeled examples, i.e. $m_L = \{100, 600, 1000, 3000\}$ following the literature used for comparative results. The unsupervised regularization batches have been formed choosing $b_L=16$ and $b_U=112$.  

\begin{figure}[h]
	\begin{center}
		\centerline{\includegraphics[width=\columnwidth,trim={0.2cm 0.2cm 0.2cm 0.2cm},clip]{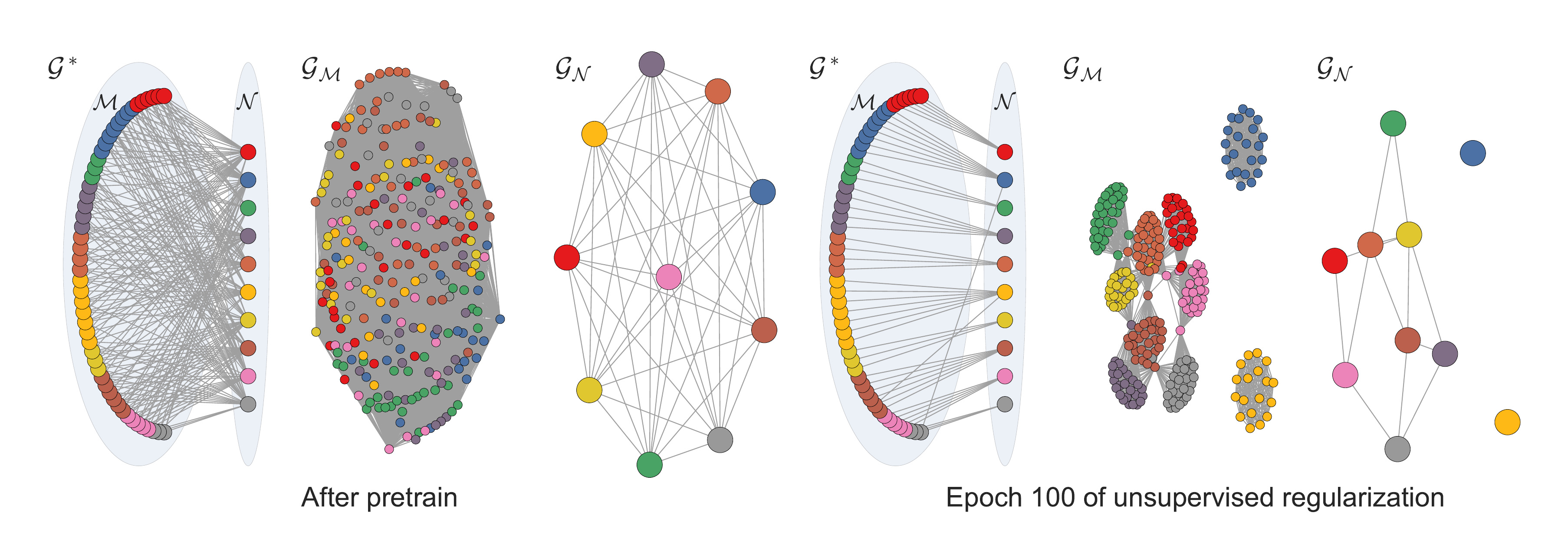}}
		\caption{Visualizations of the graphs $\mathcal{G}^*$, $\mathcal{G}_\mathcal{M}$ and $\mathcal{G}_\mathcal{N}$ for randomly chosen 250 test examples from MNIST for $m_L=100$ case. As implied in (\ref{relation}) when we regularize $\boldsymbol{N}$ to become the identity matrix: i) $\mathcal{G}^*$ becomes a biregular graph, ii) $\mathcal{G}_\mathcal{M}$ turns into a disconnected graph of $n$ $\nicefrac{m}{n}$-regular subgraphs, iii) $\mathcal{G}_\mathcal{N}$ turns into a disconnected graph of $n$ nodes having self-loops only (self-loops are not displayed for the sake of clarity), and ultimately iv) $\boldsymbol{B}$ becomes the optimal embedding that we are looking for. Color codes denote the ground-truths for the examples. This figure is best viewed in color.}
		\label{fig:img_graph}
	\end{center}
\end{figure}

\begin{figure}[h]
	\begin{center}
		\centerline{\includegraphics[width=\columnwidth,trim={0.2cm 0.2cm 0.2cm 0.2cm},clip]{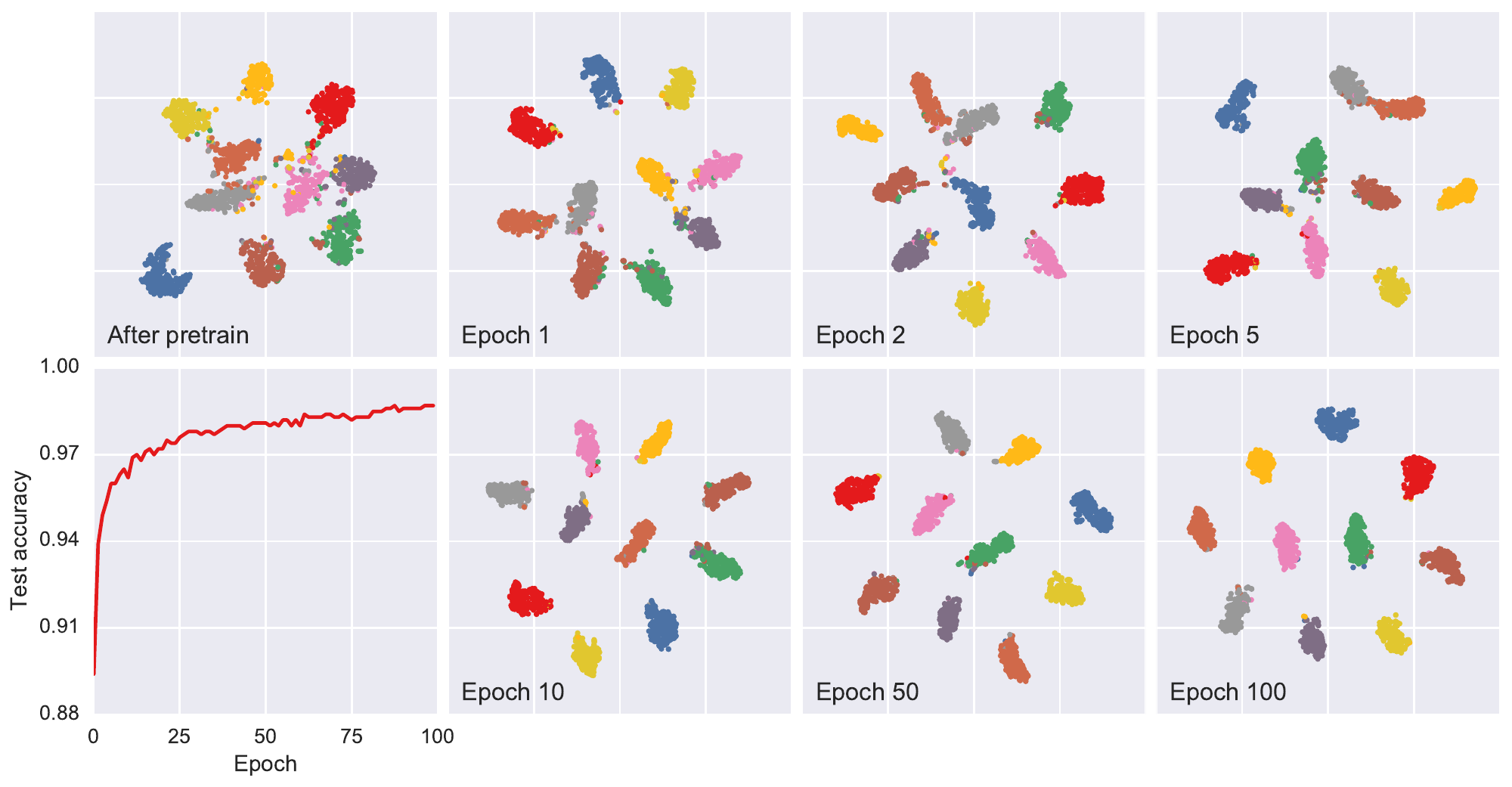}}
		\caption{t-SNE visualization of the embedding spaces inferred by $\boldsymbol{B}$ for randomly chosen 2000 test examples for $m_L=100$ case. Color codes denote the ground-truths for the examples. Note the separation of clusters from epoch 0 (right after supervised pretraining) to epoch 100 of the unsupervised training. For reference, accuracy for the entire test set is also plotted with respect to the unsupervised training epochs. This figure is best viewed in color.}
		\label{fig:img_tsne}
	\end{center}
\end{figure}

Figure \ref{fig:img_graph} visualizes the realization of the graph-based approach described in this paper using real predictions for the MNIST dataset. After the supervised pretraining, in the bipartite graph $\mathcal{G}^*$ (defined by $\boldsymbol{B}$), most of the examples are connected to multiple output nodes at the same time. In fact, the graph between the examples $\mathcal{G}_\mathcal{M}$ (inferred by $\boldsymbol{M}=\boldsymbol{B}\boldsymbol{B}^T$) looks like a sea of edges. However, thanks to pretraining, some of these edges are actually quite close to the numerical probability value of 1. Through an easier regularization objective, which is defined on the graph between the output nodes $\mathcal{G}_\mathcal{N}$ (inferred by $\boldsymbol{N}=\boldsymbol{B}^T\boldsymbol{B}$), stronger edges are implicitly propagated in graph $\mathcal{G}_\mathcal{M}$. In other words, as hypothesized, when $\boldsymbol{N}$ turns into the identity matrix, $\mathcal{G}^*$ closes to be a biregular graph and $\mathcal{G}_\mathcal{M}$ closes to be a disconnected graph of $n$ $\nicefrac{m}{n}$-regular subgraphs. As implied through the relation defined in (\ref{relation}), we expect $\boldsymbol{B}$ to ultimately become the optimal embedding that we are looking for. Figure \ref{fig:img_tsne} presents the t-SNE \cite{maaten2008tsne} visualizations of the embedding spaces inferred by $\boldsymbol{B}$. As clearly observed from this figure, as the unsupervised regularization exploits the unlabeled data, clusters become well-separated and simultaneously the test accuracy increases.

Table \ref{tab:mnist} summarizes the semi-supervised test accuracies observed with four different $m_L$ settings. Results of a broad range of recent existing solutions are also presented for comparison. These solutions are grouped according to their approaches to semi-supervised learning. While \cite{KingmaMRW14}, \cite{RasmusBHVR15}, \cite{MaaloeSSW16} employ autoencoder variants, \cite{MiyatoMKNI15}, \cite{MiyatoMKI17} and \cite{SalimansGZCRCC16} adopt adversarial training, and \cite{WestonRMC12} and \cite{WangFHLW17} are other graph-based methods used for the comparison. To show the baseline of the unsupervised regularization step in our framework, the performance of the network after the supervised pretraining is also given. For MNIST, GAR outperforms existing graph-based methods and all the contemporary methods other than some cutting-edge approaches that use generative models. 

It is theoretically difficult to make a fair complexity comparison between conventional graph-based algorithms and those utilizing neural networks with stochastic gradient descent on a newly proposed regularizer without access to the operational scripts by the authors of each algorithm.  Instead, we wanted to further analyze the computational overhead of the GAR technique in terms of time cost. To achieve this, the same CNN model was trained without any of the proposed regularization terms on an NVIDIA Geforce GTX 1080 graphics card where 100 training epochs were completed on average in 830 seconds. When GAR was activated, the same training process took 882 seconds on average which represents only a 5\% increase when compared to the time cost of the traditional neural network.  In order to show the scalability of the proposed technique to significantly larger datasets, we experimented on the InfiMNIST dataset which infinitely produces digit images derived from the original MNIST dataset using pseudo-random deformations and translations (where 8 million images were generated for the experiment described in this paper). Like the original MNIST dataset, we updated the weights using 60,000 pseudo-generated examples per epoch and continued the training until all 8 million examples were introduced to the model. While the computational overhead of the GAR technique remained the same (as we still used the same number of 60,000 examples to update the weights per epoch), introducing new examples to the model during the unsupervised regularization portion of the training further improved the performance on the test set as shown in Table \ref{tab:mnist8m}.  

\begin{table}[t]\centering
\ra{1.2}
\caption{Benchmark results of the semi-supervised test errors on MNIST for few labeled samples, i.e. 100, 600, 1000, 3000. Results of a broad range of recent existing solutions are also presented for comparison. The last row demonstrates the benchmark scores of the proposed framework in this article.}

\resizebox{\columnwidth}{!} {
\begin{tabular}{@{}lrrrr@{}}\toprule
& $m_L=100$& $m_L=600$& $m_L=1000$& $m_L=3000$\\
\midrule
M1+M2\cite{KingmaMRW14} 			& $3.33\%(\pm 0.14)$	& $2.59\%(\pm 0.05)$	& $2.40\%(\pm 0.02)$	& $2.18\%(\pm 0.04)$\\
Ladder Network\cite{RasmusBHVR15} 	& $1.06\%(\pm 0.37)$	& -						& $0.84\%(\pm 0.08)$	& -\\
AGDM\cite{MaaloeSSW16} 				& $0.96\%(\pm 0.02)$	& -						& -						& -\\
\midrule
VAT\cite{MiyatoMKNI15} 				& $2.12\%\hskip 3.35em$				& $1.39\%\hskip 3.35em$	& $1.36\%\hskip 3.35em$	& $1.25\%\hskip 3.35em$\\
Extended VAT\cite{MiyatoMKI17}		& $1.36\%\hskip 3.35em$				& -						& $1.27\%\hskip 3.35em$
& -\\
Improved GAN\cite{SalimansGZCRCC16}	& $0.93\%(\pm 0.07)$	& -			& -						& -\\
\midrule
EmbedCNN\cite{WestonRMC12} 			& $7.75\%\hskip 3.35em$				& $3.82\%\hskip 3.35em$	& $2.73\%\hskip 3.35em$	& $2.07\%\hskip 3.35em$\\
HAGR\cite{WangFHLW17} 				& $11.34\%(\pm1.23)$				& -	& -	& -\\
Pretraining (Baseline) 				& $14.82\%(\pm 1.03)$	& $4.16\%(\pm 0.26)$	& $3.25\%(\pm 0.11)$	& $1.73\%(\pm 0.08)$\\
\textbf{Pretraining	+ GAR} 		& $1.56\%(\pm 0.09)$	& $1.15\%(\pm 0.07)$	& $1.10\%(\pm 0.07)$	& $0.93\%(\pm 0.05)$\\
\bottomrule
\label{tab:mnist}
\end{tabular}
}
\end{table}

\begin{table}[t]\centering
	\ra{1.2}
	\caption{Benchmark results of the semi-supervised test errors on InfiMNIST for few labeled samples, i.e. 100. The last row demonstrates the benchmark scores of the proposed framework in this article.}
		\begin{tabular}{@{}lr@{}}\toprule
			& $m_L=100$\\
			\midrule
			HAGR\cite{WangFHLW17} 				& $8.64\%(\pm0.70)$ \\
			\textbf{Pretraining	+ GAR} 		    & $1.00\%(\pm 0.09)$\\
			\bottomrule
			\label{tab:mnist8m}
		\end{tabular}
\end{table}

\subsection{SVHN and NORB}

SVHN and NORB datasets are both used frequently in recent literature for semi-supervised classification benchmarks. Either dataset represents a significant jump in difficulty for classification when compared to the MNIST dataset. Table \ref{tab:svhn_norb} summarizes the semi-supervised test accuracies observed on SVHN and NORB. For SVHN experiments, 1000 labeled examples have been chosen among 73257 training examples. Two experiments are conducted where the SVHN \textit{extra} set (an additional training set including 531131 more samples) is either omitted from the unsupervised training or not. The same batch ratio has been used in both experiments as $b_L=96, b_U=32$. On the other hand, for NORB experiments, 2 different settings have been used for the number of labeled examples, i.e. $m_L = \{300, 1000\}$ with the same batch ratio selected as $b_L=32, b_U=96$. Both results are included for comparative purposes. Deep generative approaches have gained popularity especially over the last two years for semi-supervised learning and achieved superior performance for the problems in this setting in spite of the difficulties in their training. GAR is a low-cost and efficient competitor for generative approaches and achieves comparable performance with state-of-the-art models. Most importantly, GAR is open to further improvements with standard data enhancement techniques such as augmentation, ensemble learning.

\begin{table}[b]\centering
	\ra{1.2}
	\caption{Benchmark results of the semi-supervised test errors on SVHN and NORB. Results of a broad range of most recent existing solutions are also presented for comparison. The last row demonstrates the benchmarks for the proposed framework in this article.}
	
	\resizebox{\columnwidth}{!} {
		\begin{tabular}{@{}lrrcrr@{}}\toprule
			
& \multicolumn{2}{c}{SVHN} & \phantom{ab}& \multicolumn{2}{c}{NORB} \\ \cmidrule{2-3} \cmidrule{5-6}
& \multicolumn{1}{r}{$m_L=1000^\dagger$} & \multicolumn{1}{r}{$m_L=1000$} && \multicolumn{1}{r}{$m_L=300$} & \multicolumn{1}{r}{$m_L=1000$}\\ \midrule

M1+TSVM\cite{KingmaMRW14} 			& $55.33\%(\pm 0.11)$	& -	&& -	& $18.79\%(\pm 0.05)$\\
M1+M2\cite{KingmaMRW14} 			& $36.02\%(\pm 0.10)$	& -	&& -	& - \\
SGDM\cite{MaaloeSSW16} 				& $29.82\%\hskip 3.35em$& $16.61\%(\pm 0.24)$	&& -					& $9.40\%(\pm 0.04)$\\
\midrule
VAT\cite{MiyatoMKNI15} 				& $24.63\%\hskip 3.35em$& -	&& -	& $9.88\%\hskip 3.35em$\\
Extended VAT\cite{MiyatoMKI17}		& $5.77\%\hskip 3.35em$& -	&& -	& -						\\
Extended VAT+EntMin\cite{MiyatoMKI17}& $4.28\%\hskip 3.35em$& -	&& -	& -						\\

Improved GAN\cite{SalimansGZCRCC16}	& $8.11\%(\pm 1.30)$    & - && - 	& - 					\\ 
ALI\cite{DumoulinBPLAMC16}			& - & $7.42\%(\pm 0.65)$	&& - 	& - 					\\ 
\midrule
$\Pi$ Model \cite{LaineA16} 		& $5.43\%(\pm 0.25)$    & - && - 	& - 					\\
\midrule
Pretraining (Baseline) 				& $19.11\%(\pm 1.09)$	& $19.11\%(\pm 1.09)$	&& $17.93\%(\pm 1.07)$	& $10.22\%(\pm 1.00)$\\
\textbf{Pretraining	+ GAR} 			& $8.67\%(\pm 0.65)$	& $6.98\%(\pm 0.82)$	&& $12.19\%(\pm 1.46)$	& $7.10\%(\pm 0.57)$\\
\bottomrule
\end{tabular}
}
\raggedright \tiny $ ^\dagger$ This column presents the results obtained when SVHN \textit{extra} set is omitted from the unsupervised training. Unless otherwise specified, reported results for other approaches are assumed to represent this scenario. 
\label{tab:svhn_norb}
\end{table}

\section{Conclusion}
In this paper, we proposed a novel graph-based framework considering adaptive adjacency of the examples, $\boldsymbol{M}$, which is inferred using the predictions of a neural network model. When well-constrained, adaptive adjacency approach contributes to improved accuracy results and automatically yields that the predictions of the network become the optimal embedding without requiring any additional step such as applying Laplacian Eigenmaps. We satisfied these constraints by defining a regularization over the adjacency of the output nodes, $\boldsymbol{N}$, which is also inferred using the predictions of the network. Such regularization helped us devise an efficient and scalable framework that is natural to the operation of neural networks. Through this low-cost and easy-to-train framework, we obtained comparable performance with state-of-the-art generative approaches for semi-supervised learning.

\newpage

\bibliography{bibliography}
\bibliographystyle{apalike}

\end{document}



\maketitle
\appendix
\section{Appendices}
\subsection{Datasets used in the experiments}

Table \ref{tab:datasets} summarizes the properties of the datasets used in the experiments. The following preprocessing steps have been applied to these datasets.
\begin{itemize}
	\item \textbf{MNIST}: Images were normalized by 255. 
	\item \textbf{SVHN}: We applied sample-wise centering and normalization, i.e. we set each sample mean to 0 and divided each input by its standard deviation.
	\item \textbf{NORM}: Following \cite{MaaloeSSW16}, images were downsampled to $32 \times 32$. We added uniform noise between 0 and 1 to each pixel value. First, we normalized the NORB dataset by 256, then applied both sample-wise centering and normalization and feature-wise centering and normalization.
\end{itemize}	 

\begin{table}[h]\centering
	\ra{1.2}
	\caption{Datasets used in the experiments.}
	\resizebox{\columnwidth}{!} {
		\begin{tabular}{@{}lllll@{}}\toprule
					& Data type & Number of examples & Dimension 				& Number of classes \\
			\midrule
			MNIST 	&Image: Hand-written digits & Train: 60000, Test: 10000		& $1\times28\times28$		& 10				\\
			SVHN	&Image: Street-view digits & Train: 73257, Test: 26032, Extra: 531131		& $3\times32\times32$		& 10				\\
			NORB	&Image: Objects& Train: 24300, Test: 24300		& $2\times96\times96$		& 5				\\
			\bottomrule
			\label{tab:datasets}
		\end{tabular}
	}
\end{table}

\subsection{Models used in the experiments}

Table \ref{tab:models} summarizes all models used in the experiments. Reported MNIST results have been obtained using the model named \textit{6-layer CNN}, whereas SVHN results were obtained on \textit{9-layer CNN-2} model and NORB results were obtained on \textit{9-layer CNN} model. The results obtained on different models are also presented in the following sections to show the effect of the chosen model to the test accuracy. The ReLU activation function has been used for all models. For both supervised pretraining and unsupervised regularization, models were trained using stochastic gradient descent with following settings: $lr=0.01$, $decay=1e^{-6}$, $momentum=0.95$ with Nesterov updates.
 
\begin{table}[h]\centering
	\ra{1.2}
	\caption{Specifications of two CNN models used in the experiments.}
	\resizebox{\columnwidth}{!} {
		\begin{tabular}{@{}llllll@{}}\toprule
			Model name & Specification & & & &\\
			\midrule
4-layer MLP & FC 2048 - Drop(0.5) & FC 2048 - Drop(0.5) & FC 2048 - Drop(0.5) & FC 10 &\\
6-layer CNN & 2*32x3x3 - MP2x2 - Drop(0.2) & 2*64x3x3 - MP2x2 - Drop(0.3) & FC 2048 - Drop(0.5) & FC 10 &\\	
9-layer CNN & 2*32x3x3 - MP2x2 - Drop(0.2) & 2*64x3x3 - MP2x2 - Drop(0.3) & 3*128x3x3 - MP2x2 - Drop(0.4) & FC 2048 - Drop(0.5) & FC 10\\	
9-layer CNN-2 & 2*64x3x3 - MP2x2 - Drop(0.2) & 2*128x3x3 - MP2x2 - Drop(0.3) & 3*256x3x3 - MP2x2 - Drop(0.4) & FC 2048 - Drop(0.5) & FC 10\\
	 	
			\bottomrule
			\label{tab:models}
		\end{tabular}
	}
\end{table}

\subsection{Effects of hyperparameters}

\subsubsection{Effect of the chosen model}

\begin{figure}[h]
	\begin{center}
		\centerline{\includegraphics[width=\columnwidth,trim={0.2cm 0.2cm 0.2cm 0.2cm},clip]{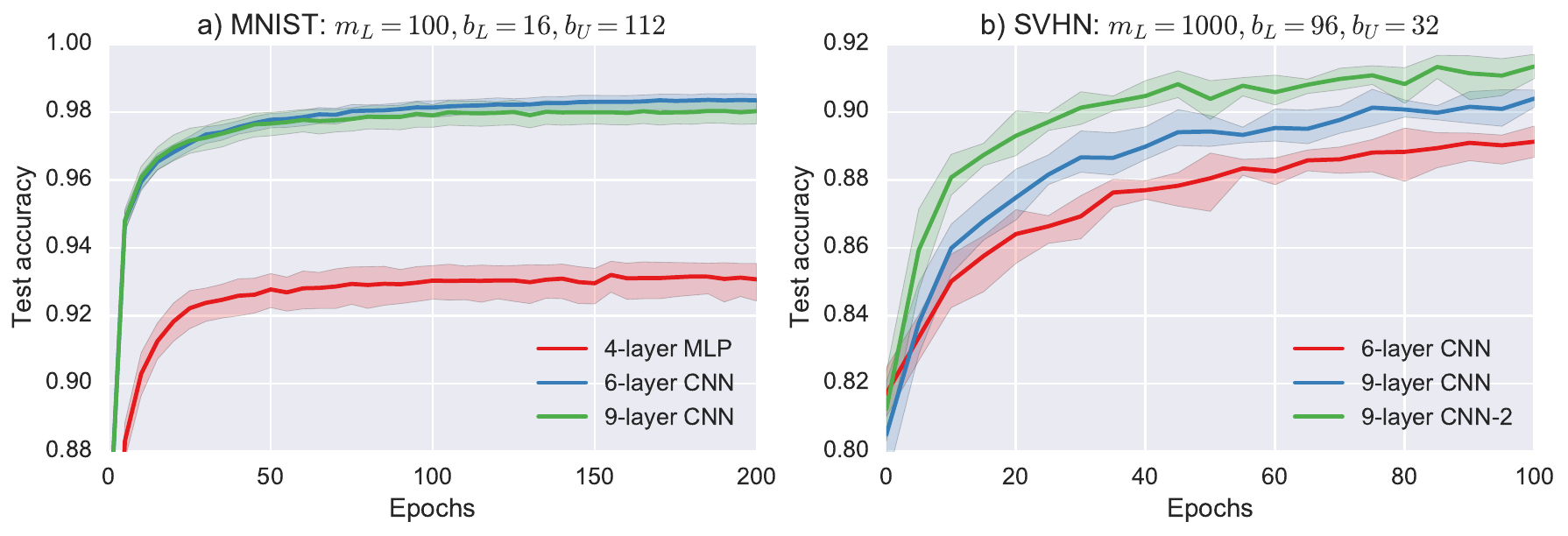}}
		\caption{Test accuracies obtained using different models with respect to unsupervised training epochs. a) MNIST results are given for the settings that $m_L=100$ and $b_L=16, b_U=112$ and b) SVHN results are given for the settings that $m_L=1000$ and $b_L=96, b_U=32$. Model specifications are given in Table \ref{tab:models}.} 
		\label{fig:img_appdx_model}
	\end{center}
\end{figure}

Figure \ref{fig:img_appdx_model} presents the test accuracy curves with respect to the unsupervised training epochs obtained using different models. The proposed unsupervised regularization improves the test accuracy in all models. However, the best case depends on the chosen model specifications. 

\subsubsection{Effect of $\nicefrac{b_L}{b_U}$ ratio and applying dropout during unsupervised regularization}

\begin{figure}[h]
	\begin{center}
		\centerline{\includegraphics[width=\columnwidth,trim={0.2cm 0.2cm 0.2cm 0.2cm},clip]{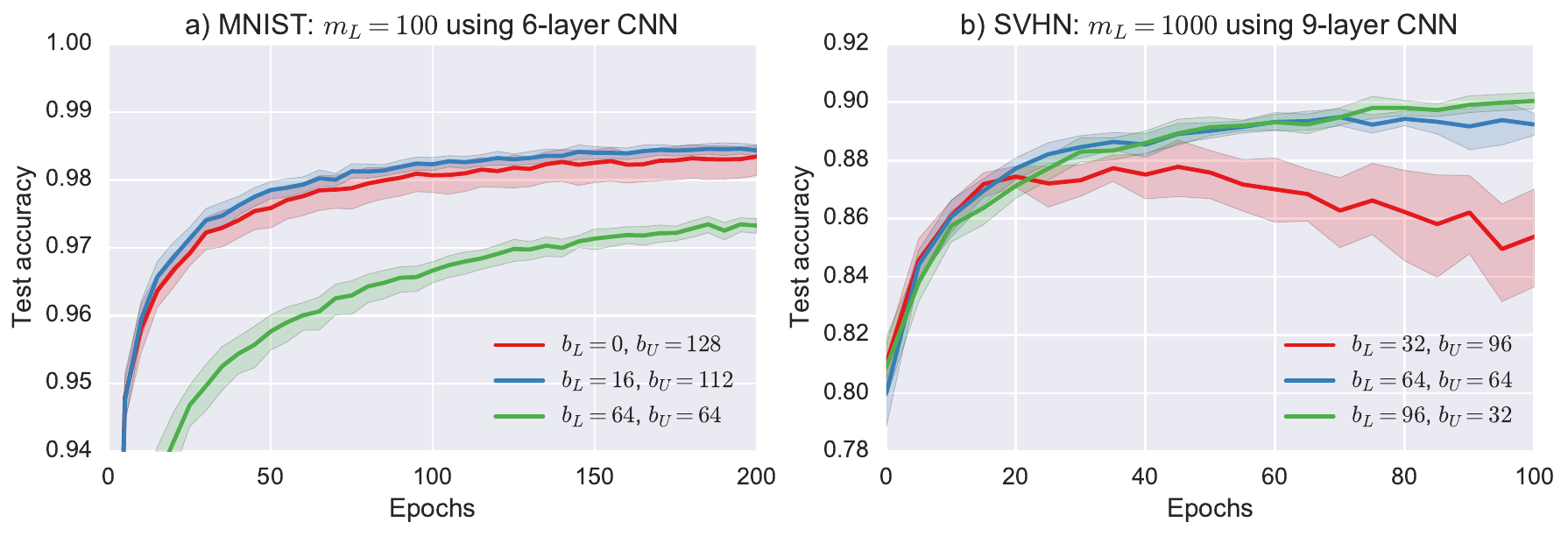}}
		\caption{The effect of $\nicefrac{b_L}{b_U}$ ratio for MNIST and SVHN datasets.}
		\label{fig:img_appdx_batch_ratio}
	\end{center}
\end{figure}

The labeled/unlabeled data ratio of unsupervised training batches is the most critical hyperparameter of the proposed regularization. Figure \ref{fig:img_appdx_batch_ratio} visualizes the effect of this ratio for MNIST and SVHN datasets. These two datasets have different characteristics. MNIST dataset has a lower variance among its samples with respect to SVHN. As a result, even when the labeled examples introduced to the network during the supervised pretraining are not blended to the unsupervised training batches, i.e. $b_L = 0$, this does not affect the performance dramatically. However, for SVHN dataset, reducing the $b_L$ proportion of the unsupervised training batches significantly affects the accuracy and further decrease of $b_L$ reduces the stability of the regularization.

\begin{figure}[h]
	\begin{center}
		\centerline{\includegraphics[width=\columnwidth,trim={0.2cm 0.2cm 0.2cm 0.2cm},clip]{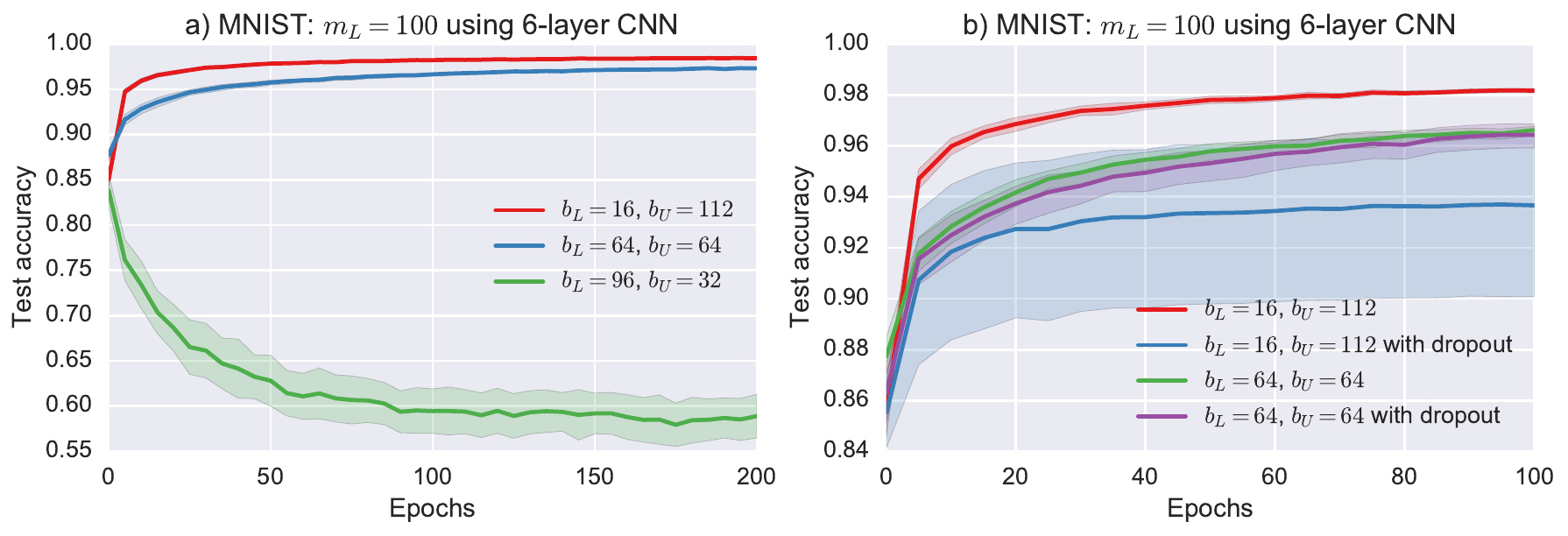}}
		\caption{The effects of a) choosing $b_L \approx m_L$ and b) applying dropout during the unsupervised training on MNIST dataset.}
		\label{fig:img_appdx_bad_hypers}
	\end{center}
\end{figure}

One can also observe another phenomenon through the MNIST results in Figure \ref{fig:img_appdx_batch_ratio}. That is, as $b_L$ approaches to $m_L$, the generalization of the model reduces. This effect can be better observed in Figure \ref{fig:img_appdx_bad_hypers} including a further step, i.e. $b_L=96$ when $m_L=100$. Since the same examples start to dominate the batches of unsupervised regularization, overfitting occurs and ultimately test accuracy significantly reduces. Figure \ref{fig:img_appdx_bad_hypers} also presents the effect of applying dropout during the unsupervised regularization. Dropping out the weights the during unsupervised training dramatically affects the accuracy. This effect is more obvious when ${b_L}$ is smaller. Hence, for the experiments, we removed the dropouts from the models specified in Table \ref{tab:models} during the unsupervised training and applied the following strategy to decide on the $\nicefrac{b_L}{b_U}$ ratio: $b_L$ is approximately assigned as one tenth of the number of all labeled examples $m_L$, i.e. $b_L \approx \nicefrac{m_L}{10}$, and then $b_U$ is chosen to complement the batch size up to 128, i.e. $b_U=128-b_L$.  

\subsubsection{Effect of regularization coefficients $c_\alpha$, $c_\beta$ and $c_F$ }

\begin{figure}[h]
	\begin{center}
		\centerline{\includegraphics[width=\columnwidth,trim={0.2cm 0.2cm 0.2cm 0.2cm},clip]{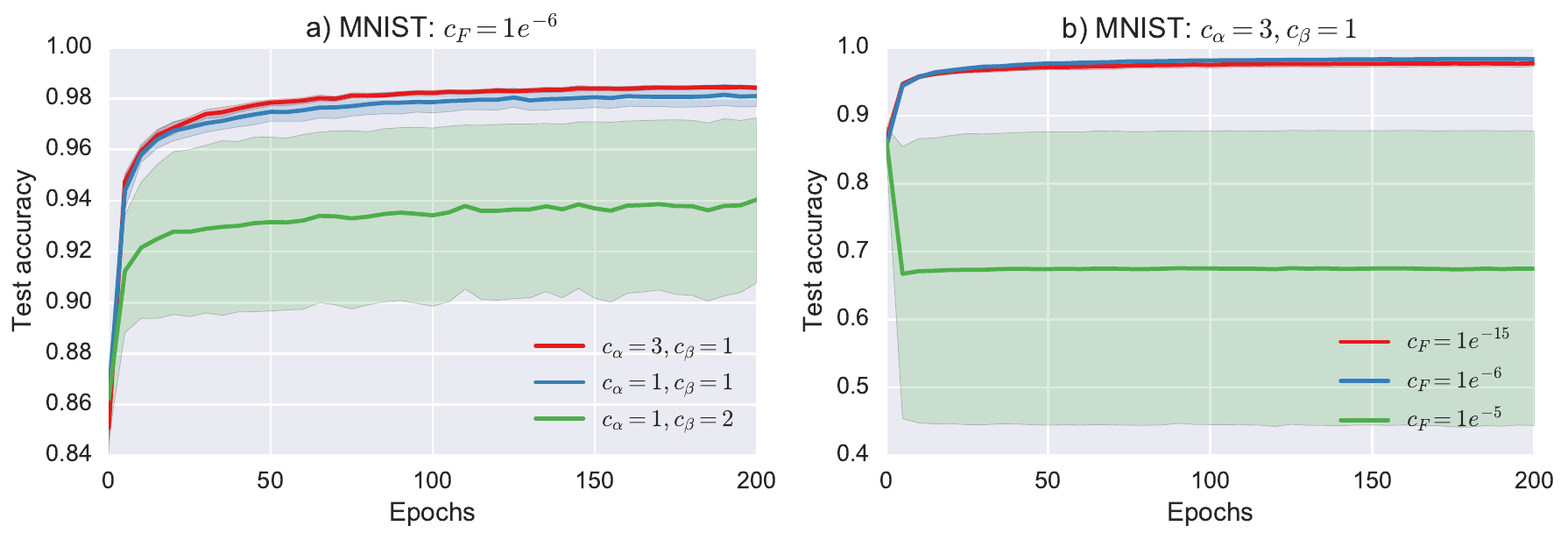}}
		\caption{The effects of a) $\nicefrac{c_\alpha}{c_\beta}$ and b) $c_F$ on MNIST dataset. Shaded green regions indicate the instability of the model when using the specified settings as the obtained accuracy varies drastically per epoch. }
		\label{fig:img_reg_coeffs}
	\end{center}
\end{figure}

The effects of regularization coefficients are presented in Figure \ref{fig:img_reg_coeffs} for MNIST dataset. Part (a) of the figure visualizes the case when $c_F$ is held constant, but the ratio of $\nicefrac{c_\alpha}{c_\beta}$ changes. And part (b) of the figure illustrates the case when $\nicefrac{c_\alpha}{c_\beta}$ is held constant, but $c_F$ changes. We can say that as long as $c_\alpha \ge c_\beta$, the ratio of $\nicefrac{c_\alpha}{c_\beta}$ does not affect the accuracy significantly. Furthermore, the value of $c_F$ is not so critical (close performances both with $c_F=1e^{-6}$ and $c_F=1e^{-15}$) unless it is too big to distort the regularization. Therefore, we can say that the proposed unsupervised regularization term is considerably robust with respect to the coefficients $c_\alpha$, $c_\beta$ and $c_F$. This can also be seen through the fact that we have applied the same coefficients for the experiments of all three datasets.

\subsection{More on activity regularization}
Recall that 
\begin{equation}
\boldsymbol{N}:=\boldsymbol{B}^T\boldsymbol{B}=
\begin{bmatrix}
\sum\limits^{m}{B_{i1}B_{i1}} & \sum\limits^{m}{B_{i1}B_{i2}} & \dots  & \sum\limits^{m}{B_{i1}B_{in}} \\
\sum\limits^{m}{B_{i2}B_{i1}} & \sum\limits^{m}{B_{i2}B_{i2}} & \dots  & \sum\limits^{m}{B_{i2}B_{in}} \\
\vdots & \vdots & \ddots & \vdots \\
\sum\limits^{m}{B_{in}B_{i1}} & \sum\limits^{m}{B_{in}B_{i2}} & \dots  & \sum\limits^{m}{B_{in}B_{in}} 
\end{bmatrix}
\end{equation}

then we can rewrite $\alpha(\boldsymbol{B}\big)$ in terms of $\boldsymbol{B}$ only

\begin{equation}
\alpha\big(\boldsymbol{B}\big) =\frac{\sum\limits_{i \ne j}^n{N_{ij}}}{(n-1)\sum\limits_{i = j}^n{N_{ij}}}=
\frac{2\sum\limits_{i=1}^{n-1} \sum\limits_{j=i+1}^n N_{ij}}{(n-1)\sum\limits_{i=1}^n{N_{ii}}}=
\frac{2\sum\limits_{i=1}^{n-1} \sum\limits_{j=i+1}^n \sum\limits_{k=1}^{m}B_{ki}B_{kj}}{(n-1)\sum\limits_{i=1}^n{\sum\limits_{k=1}^{m}B_{ki}B_{ki}}}
\end{equation}

Also recall that $\boldsymbol{v}$ represents the diagonal entries of $\boldsymbol{N}$ such that $\boldsymbol{v}:=\begin{bmatrix}N_{11} & N_{22} & \dots  & N_{nn}\end{bmatrix}$ and

\begin{equation}
\boldsymbol{V}:=\boldsymbol{v}^T\boldsymbol{v}=
\begin{bmatrix}
N_{11}N_{11} & N_{11}N_{22} & \dots  & N_{11}N_{nn} \\
N_{22}N_{11} & N_{22}N_{22} & \dots  & N_{22}N_{nn} \\
\vdots & \vdots & \ddots & \vdots \\
N_{nn}N_{11} & N_{nn}N_{22} & \dots  & N_{nn}N_{nn}
\end{bmatrix}
=\begin{tiny}
\begin{bmatrix}
\sum\limits^{m}{B_{i1}^2}\sum\limits^{m}{B_{i1}^2} & \sum\limits^{m}{B_{i1}^2}\sum\limits^{m}{B_{i2}^2}& \dots  & \sum\limits^{m}{B_{i1}^2}\sum\limits^{m}{B_{in}^2} \\
\sum\limits^{m}{B_{i2}^2}\sum\limits^{m}{B_{i1}^2} & \sum\limits^{m}{B_{i2}^2}\sum\limits^{m}{B_{i2}^2} & \dots  & \sum\limits^{m}{B_{i2}^2}\sum\limits^{m}{B_{in}^2} \\
\vdots & \vdots & \ddots & \vdots \\
\sum\limits^{m}{B_{in}^2}\sum\limits^{m}{B_{i1}^2} & \sum\limits^{m}{B_{in}^2}\sum\limits^{m}{B_{i2}^2} & \dots  & \sum\limits^{m}{B_{in}^2}\sum\limits^{m}{B_{in}^2}
\end{bmatrix}
\end{tiny}
\end{equation}

$1-\beta(\boldsymbol{B}\big)$ can be written in terms of $\boldsymbol{N}$ as follows:

\begin{equation}
1-  \beta\big(\boldsymbol{B}\big) = 1 - \frac{\sum\limits_{i \ne j}{V_{ij}}}{(n-1)\sum\limits_{i = j}{V_{ij}}} = 
\frac{\sum\limits_{i=1}^{n-1} \sum\limits_{j=i+1}^n (N_{ii}-N_{jj})^2}{(n-1)\sum\limits_{i=1}^n{N_{ii}^2}}
\end{equation}

If we further replace $\boldsymbol{N}$ with $\boldsymbol{B}$ then 
\begin{equation}
1-  \beta\big(\boldsymbol{B}\big) = \frac{\sum\limits_{i=1}^{n-1} \sum\limits_{j=i+1}^n \bigg(\sum\limits_{k=1}^{m}B_{ki}B_{ki}-\sum\limits_{k=1}^{m}B_{kj}B_{kj}\bigg)^2}{(n-1)\sum\limits_{i=1}^n{\sum\limits_{k=1}^{m}(B_{ki}B_{ki})^2}}
\end{equation}

Recall that the proposed unsupervised loss is 

\begin{equation}
\label{unsupervised_objective}
\mathcal{U}\big(\boldsymbol{B}\big)= c_{\alpha}\alpha\big(\boldsymbol{B}\big) + c_{\beta}\big(1-\beta\big(\boldsymbol{B}\big)\big) + c_F||\boldsymbol{B}||^2_F
\end{equation}

then the overall unsupervised loss can be written in terms of $\boldsymbol{B}$ as follows:

\begin{equation}
\mathcal{U}\big(\boldsymbol{B}\big) = c_\alpha\begin{pmatrix}
\frac{2\sum\limits_{i=1}^{n-1} \sum\limits_{j=i+1}^n \sum\limits_{k=1}^{m}B_{ki}B_{kj}}{(n-1)\sum\limits_{i=1}^n{\sum\limits_{k=1}^{m}B_{ki}^2}}\end{pmatrix}+c_\beta\begin{pmatrix}\frac{\sum\limits_{i=1}^{n-1} \sum\limits_{j=i+1}^n \bigg(\sum\limits_{k=1}^{m}B_{ki}^2-\sum\limits_{k=1}^{m}B_{kj}^2\bigg)^2}{(n-1)\sum\limits_{i=1}^n{\sum\limits_{k=1}^{m}B_{ki}^4}}\end{pmatrix}+c_F\begin{pmatrix}\sum\limits_{i=1}^n \sum\limits_{k=1}^mB_{ki}^2\end{pmatrix}
\end{equation}

where $m$ is the number of examples, $n$ is the number of output nodes and 
\begin{equation}
\boldsymbol{B} = g\big(\boldsymbol{X}\big) = \max{\bigg(\boldsymbol{0}, \big(\boldsymbol{Y}^{(L-1)}\boldsymbol{W}^{(L)} + \boldsymbol{b}^{(L)}\big)\bigg)}
\end{equation}

\bibliography{bibliography}
\bibliographystyle{apalike}